\documentclass[twocolumn,10pt]{article} 

\usepackage[square,numbers,sort&compress,comma]{natbib}

\usepackage{amsmath}
\usepackage{amssymb}
\usepackage{caption}
\usepackage{graphicx}
\usepackage{latexsym}
\usepackage{times}
\usepackage{color} 

\usepackage{subfigure} 
\usepackage{mlmath}

\topmargin - 12pt 
\renewenvironment{abstract}%
              {
               \small
               {\bfseries \abstractname}
               \par
               \vspace{10pt}
              }

\renewcommand\abstractname{Abstract}

\newcommand{\nomenclature}
              [1]
              {
               \bgroup
               \flushleft
               \small\bf
               #1
               \par
               \egroup
              }

\renewcommand{\section}
              [1]
              {
               \bgroup
               \flushleft
               \small\bf
               \stepcounter{section}
               \arabic{section}. #1
               \par
               \egroup
              }

\renewcommand{\subsection}
              [1]
              {
               \bgroup
               \flushleft
               \small\em
               \stepcounter{subsection}
               \arabic{section}.
               \arabic{subsection}. #1
               \par
               \egroup
              }

\renewcommand{\subsubsection}
              [1]
              {
               \bgroup
               \flushleft
               \small\em
               \stepcounter{subsubsection}
               \arabic{section}.
               \arabic{subsection}.
               \arabic{subsubsection}. #1
               \par
               \egroup
              }

  \newcommand{\acknowledgement}
              [1]
              {
               \bgroup
               \flushleft
               \small\bf
               #1
               \par
               \egroup
              }

  \newcommand{\sectionbib}
              [1]
              {
               \bgroup
               \flushleft
               \small\bf
               #1
               \par
               \egroup
              }

\usepackage{xcolor}  

\begin{document}

\title{\LARGE A deep learning-based model reduction (DeePMR) method for simplifying chemical kinetics}

\author{{\large Zhiwei Wang$^{a,b}$, Yaoyu Zhang$^{a,b,c}$, Enhan Zhao$^{g}$, Yiguang Ju$^{d}$, Weinan E$^{e,f}$, }\\[10pt]
{\large Zhi-Qin John Xu$^{a,b,*}$, Tianhan Zhang$^{f,*}$}\\[10pt]
        {\footnotesize \em $^a$Institute of Natural Sciences, School of Mathematical Sciences, Shanghai Jiao Tong University, Shanghai, 200240, China }\\[-5pt]
        {\footnotesize \em $^b$MOE-LSC and Qing Yuan Research Institute, Shanghai Jiao Tong University, Shanghai, 200240, China}\\[-5pt]
        {\footnotesize \em $^c$Shanghai Center for Brain Science and Brain-Inspired Technology, Shanghai, 200240, China}\\[-5pt]
        {\footnotesize \em $^d$Department of Mechanical and Aerospace Engineering, Princeton University, NJ, 08540, US}\\[-5pt]
        {\footnotesize \em $^e$School of Mathematical Sciences, Peking University, Beijing, 100871, China}\\[-5pt]
        {\footnotesize \em $^f$AI for Science Institute, Beijing, 100080, China}\\[-5pt]
        {\footnotesize \em $^g$School of Electronics Engineering and Computer Science, Peking University, Beijing, 100871, China}\\[-5pt]
        {\footnotesize \em $^*$Corresponding authors: xuzhiqin@sjtu.edu.cn, tianhanz@princeton.edu}\\[-5pt]
}
\date{}


\small
\baselineskip 10pt


\twocolumn[\begin{@twocolumnfalse}
\vspace{50pt}
\maketitle
\vspace{40pt}
\rule{\textwidth}{0.5pt}
\begin{abstract} 
    A deep learning-based model reduction (DeePMR) method for simplifying chemical kinetics is proposed and validated using high-temperature auto-ignitions, perfectly stirred reactors (PSR), and one-dimensional freely propagating flames of n-heptane/air mixtures. The mechanism reduction is modeled as an optimization problem on Boolean space, where a Boolean vector, each entry corresponding to a species, represents a reduced mechanism. The optimization goal is to minimize the reduced mechanism size given the error tolerance of a group of pre-selected benchmark quantities. The key idea of the DeePMR is to employ a deep neural network (DNN) to formulate the objective function in the optimization problem. In order to explore high dimensional Boolean space efficiently, an iterative DNN-assisted data sampling and DNN training procedure are implemented. The results show that DNN-assistance improves sampling efficiency significantly, selecting only $10^5$ samples out of $10^{34}$ possible samples for DNN to achieve sufficient accuracy. The results demonstrate the capability of the DNN to recognize key species and reasonably predict reduced mechanism performance. The well-trained DNN guarantees the optimal reduced mechanism by solving an inverse optimization problem. By comparing ignition delay times, laminar flame speeds, temperatures in PSRs, the resulting skeletal mechanism has fewer species (45 species) but the same level of accuracy as the skeletal mechanism (56 species) obtained by the Path Flux Analysis (PFA) method. In addition, the skeletal mechanism can be further reduced to 28 species if only considering atmospheric, near-stoichiometric conditions (equivalence ratio between 0.6 and 1.2). The DeePMR provides an innovative way to perform model reduction and demonstrates the great potential of data-driven methods in the combustion area.
\end{abstract}
\vspace{10pt}
\parbox{1.0\textwidth}{\footnotesize {\em Keywords:} model reduction; machine learning; deep neural network; chemical kinetics}
\rule{\textwidth}{0.5pt}
\vspace{10pt}
\end{@twocolumnfalse}] 


\clearpage

\section{Introduction} 
\begin{figure}[h!]
        \centering
        \includegraphics[width=192pt]{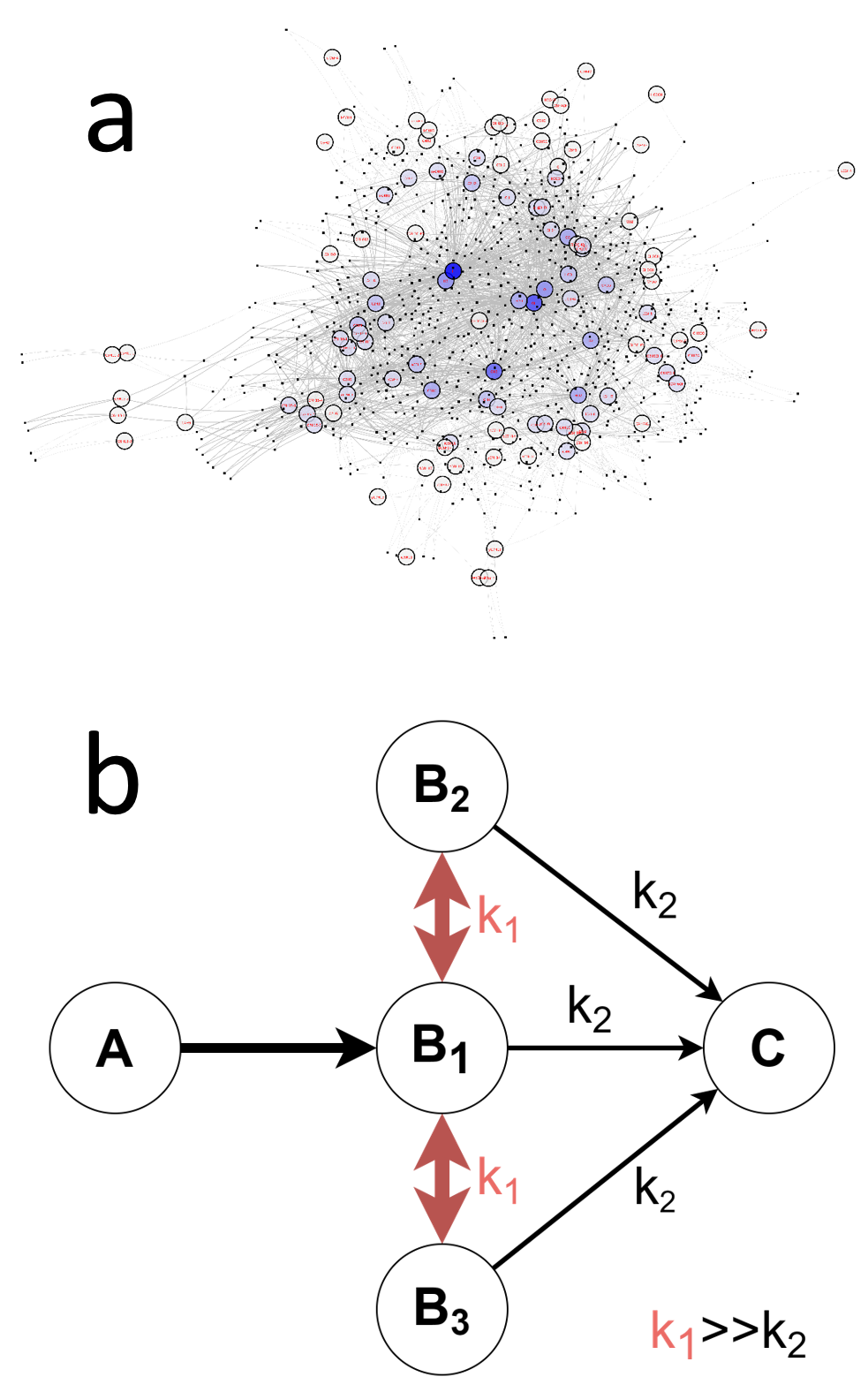}
        \caption{(a): Species-reaction graph for detailed n-heptane mechanism (116 species, 830 reactions). Circles represent species, and black dots represent reactions. The circles are dyed by the vortex degree; (b): An illustration for a reaction pathway.}
        \label{fig: sr_graph_116}
    \end{figure}

\addvspace{10pt}
    The development of detailed chemistry mechanisms of hydrocarbon fuels paves the way to realistic simulations of practical combustors \cite{Lu2009}. However, due to chemistry stiffness, the simulation of large-size detailed mechanisms become forbiddingly expensive, especially for very large-scale simulation. Figure \ref{fig: sr_graph_116}a shows a species-relation graph of the n-heptane mechanism with 116 species and 830 reactions, revealing an intrinsic complexity of the chemistry mechanism \cite{chaos2007}. It is a highly non-trivial task to remove part of the chemistry mechanism but maintain its overall accuracy for simulations in various reactors. Therefore, there is a substantial need to develop a systematic way to simplify chemistry mechanisms. 
    
    Previous researchers have proposed various model reduction methods. The first type of reduction method is by sensitivity analysis \cite{Rabitz1983, TuranyiTamas1990}. The second type is time scale analysis, including quasi-steady-state approximations (QSSA) \cite{peters1985}, partial equilibrium assumptions (PEA) \cite{smoke1991},  computational singularity perturbation (CSP) \cite{Lam1989, Lam1993, valorani2006}, rate-controlled constrained-equilibrium (RCCE) \cite{keck1990}, and lifetime analysis \cite{Lovas2000}.
    
    The third type is graph-based methods, including directed relation graph (DRG) \cite{Lu2005}, directed relation graph with error propagation (DRGEP) \cite{PEPIOTDESJARDINS2008}, path flux analysis (PFA) \cite{Sun2010}, and directed relation graph with error propagation and sensitivity analysis (DRGEPSA) \cite{Niemeyer2010}. The methods focus on maintaining the fluxes in the reaction network associated with a set of targeted species when removing species. The major difference among the DRG, DRGEP, and PFA methods is the interaction coefficient definition. First, DRG, DRGEP, and PFA use the absolute, net, summation of production and consumption rates to define the interaction coefficient for the directly related species, respectively. In addition, DRG assumes that error does not decay along the graph-search path and adopts the path with maximum error. In contrast, DRGEP assumes that the error induced by indirectly related species decays geometrically. Wu et al. \cite{Wu2020} showed that DRG might overestimate the errors in the starting species, and DRGEP tends to underestimate reduction errors since the geometric error decay may overly predict decay speed. The PFA method combines the features of DRG and DRGEP, and proposes a new scheme that considers error decay but only up to the second order. Besides the error estimation method, an important topic for graph-based methods is choosing the target set of species. Chen and Chen \cite{Chen2018} showed that with or without H radical in the target set significantly impacts flame speed prediction accuracy. Curtis et al. \cite{Curtis2015} proposed the relative importance index (RII) method to determine the target species set automatically. In summary, the interaction coefficient definition and the set of target species determine the path flux error triggered by the removal of species. 
    
    Generally speaking, it is the key assumption of graph-based methods that flux error evaluation can represent the overall performance of the reduced mechanisms. However, the validity of the current assumption heavily relies on the specific interaction coefficient definition. It is unclear when the assumption fails. For example, Sun et al. \cite{Sun2010} pointed out that DRGEP cannot identify the importance of $NO_x$ catalytic effect on ignition enhancement due to using net reaction rates in error estimation. 
    
    Figure \ref{fig: sr_graph_116}b shows another example where three graph-based methods fail to perform the reduction. Figure \ref{fig: sr_graph_116}b represents a simplified mechanism, where $A$, $B_i$, $C$ are the reactant, intermediate species, and product, respectively. The inter-conversion rates among $B_1$, $B_2$, and $B_3$ are faster than other reaction rates. Consequently, $B_i$ has similar concentrations and contributions to the target species $C$. Based on error estimation methods mentioned above, there is little difference among the impacts of $B_i$’s on the flux to the target species $C$. As a result, $B_i$ will either be removed or retained altogether. In contrast, it is readily seen that removing $B_2$ and $B_3$ will not affect the reactant’s consumption rate and product’s creation rate, which is an acceptable reduction for the original mechanism, and graph-based methods do not come up with such a solution. Due to the extreme complexity of the detailed mechanism shown in Figure \ref{fig: sr_graph_116}a, it is challenging for graph-based methods to evaluate to what extent the path flux can accurately indicate the overall performance of the reduced mechanisms and where the assumption might fail.

    On the other hand, there is a recent spike of machine learning in combustion kinetic studies. For generating detailed chemistry mechanism, Zeng et al. \cite{Zeng2020} adopted a novel DeePMD method \cite{Zhang2018h} to build a detailed methane combustion mechanism using {\it ab initio} molecular dynamics (AIMD). In terms of model reduction, Si \cite{Si2021} and Kelly \cite{Kelly2021} used machine learning algorithms to optimize the parameters in the chemistry mechanism. 
    
    This paper develops a data-driven method based on the deep neural network (DNN) to reduce species and reactions in detailed mechanisms. First, we will introduce the basic idea of constructing an optimization problem for model reduction. The mathematical formulation is straightforward, while the challenge is to design a DNN to represent the objective function. The training-sampling iteration, the DNN-assisted data sampling strategy, and the inverse optimization problem will be introduced in detail. Then, the iterative training process will be discussed, including the statistical features of the sampled data and the DNN prediction performance. Afterward, the resulting reduced mechanism (45 species) is validated by comparing the ignition delay times, laminar flame speeds, and extinction curves in PSRs. Moreover, a smaller-size reduced mechanism is achieved by narrowing the range of benchmark quantities, demonstrating the flexibility of the current method. Finally, conclusions are drawn.
    
\section{Methodology} \addvspace{10pt}
In this section, we would describe our proposed deep learning-based model reduction (DeePMR) method to find optimal reduced mechanisms, followed by the implementation details in next section.
\subsection{Optimization framework for searching optimal reduced mechanisms}
    Suppose we have a detailed mechanism with $N_{s}$ species and simplify it by removing unnecessary species. First, to represent a reduced mechanism, we employ an $N_{s}$-dimensional Boolean vector $\vx\in\{0,1\}^{N_{s}}$, each entry indicating the status of a species: true or one means remaining, false or zero removing, similar to \cite{Bhattacharjee2003}. For convenience, we use float numbers $0$ and $1$ instead of True or False to implement our algorithm. Then, we assume the performance of the reduced mechanism can be indicated by a function ${\rm Err}(\vx)$. For example, function ${\rm Err}(\vx): \{0,1\}^{N_{s}}\rightarrow \mathbb{R}^{N_{t}}$ can be the errors of the reduced mechanism $\vx$ in $N_{t}$ benchmark quantities, such as the ignition delay times and laminar flame speeds under various initial conditions, temperature in PSR with various residence times. The ultimate goal is to find such an optimal solution of $\vx$ that the error of the reduced mechanism $\vx$ is within our tolerance, and the size of the reduced mechanism $\vx$, in other words, the sparsity of the vector $\vx$, is as small as possible. The reduction procedure can be considered as a typical optimization problem:
    \begin{equation}
        \min F(\vx) := \sum_{i=1}^{N_{t}} {\rm Err}(\vx)_i + \beta ||\vx||_0.
    \end{equation}
    where ${\rm Err}(\vx)_i$ is the error of the $i$-th benchmark quantity, $\beta$ is the weight of the sparsity.

\subsection{Deep neural network modeling performance function}
    Once the objective function $F(\vx)$ is determined, the optimal solution can be achieved by optimization. However, the major difficulty is parametrizing the function ${\rm Err}(\vx)$. The function ${\rm Err}(\vx)$ needs to reflect the overall performance of a reduced mechanism. The corresponding benchmark quantity can include the ignition delay time, equilibrium temperature, laminar flame speed, and temperature in PSRs. 
    
    The current work proposes an end-to-end DNN to build up the relationship between the reduced mechanism and its accuracy in predicting benchmark quantities. The input is an $N_{s}$-dimensional Boolean vector $\vx$, meaning a reduced mechanism, and output $\vu_{\vtheta}(\vx)$ is an $N_{t}$-dimensional real vector. Each element of the output vector $\vu_{\vtheta}(\vx)$ represents the prediction of the benchmark quantities of the reduced model. The structure is shown in Figure \ref{fig:dnn_structure}a. An $L$-layer neural network is,
    \begin{align}
        \vu_{\vtheta}(\vx)& = \vW^{[L-1]} \sigma\circ(\mW^{[L-2]}\sigma\circ(\cdots ( \nonumber\\
        &\mW^{[1]} \sigma\circ(\mW^{[0]} \vx + \vb^{[0]} ) + \vb^{[1]} )\cdots) \nonumber\\
        &+\vb^{[L-2]})+\vb^{[L-1]},\nonumber
    \end{align}
    where $\mW^{[l]} \in \sR^{m_{l+1}\times m_{l}}$, $\vb^{[l]}=\sR^{m_{l+1}}$, $m_0=N_s$, $m_{L}=N_{t}$, ``$\circ$'' means entry-wise operation, $\sigma$ is ReLU. We denote the set of parameters by $\vtheta$.
    Due to the intrinsic non-linearity and high-dimension of the performance function ${\rm Err}(\vx)$, we use the DNN of three hidden layers, with 2000, 1000, 1000 nodes, respectively. In addition, each type of benchmark quantity requires a DNN. For example, three DNNs are trained to predict ignition delay time, temperature, and flame speed, respectively. The loss function is Mean Square Error (MSE), and the optimizer is Stochastic Gradient Descent (SGD) with batch size 128. 
    
    The well-trained DNN serves as the objective function in the inverse optimization problem. Here, the DNN training process is considered as the 'forward' optimization problem, which optimizes the parameters in the DNN.  The 'inverse' means finding the optimal input $\vx$ given the objective function, which is in the form of a neural network. SGD is performed to find an optimal solution $\vx$ of the objective function $F(\vx)$. It is worth noting that each element $\vx_i$ of the original optimal solution $\vx$ can be any real number, but it is preferred that $\vx_i$ is close to zero or one so that $\vx$ can be easily mapped to a reduced mechanism. Consequently, an additional Sigmoid layer is added in front of the frozen DNN. The structure of the performance function is shown in Figure \ref{fig:dnn_structure}b. 

\subsection{DNN-assisted sampling method}
\begin{figure}[h!]
        \centering
        \includegraphics[width=192pt]{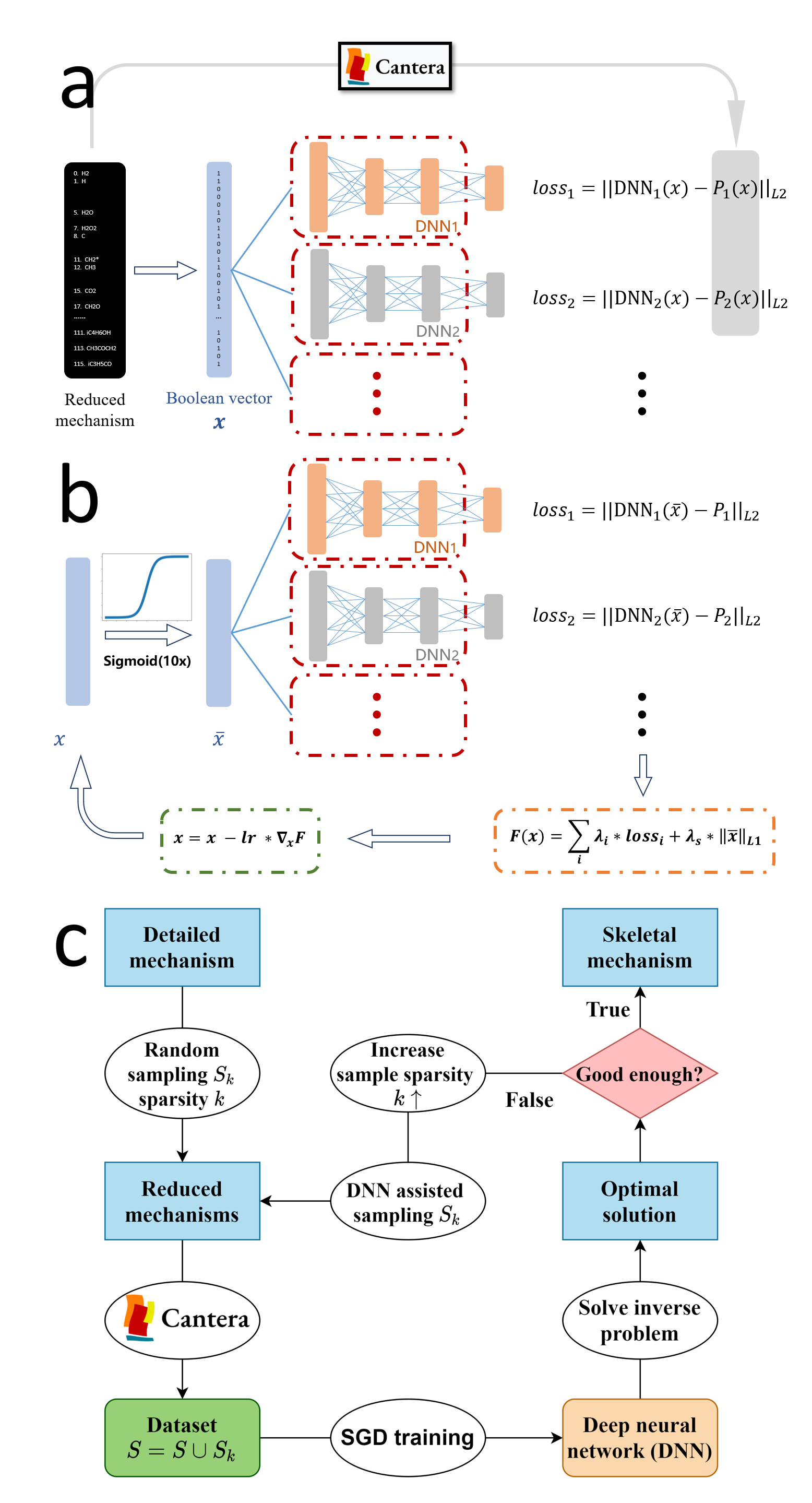}
        \caption{(a): illustration for the data labeling and DNN structure; (b): illustration for the inverse optimization and the DNN structure with an additional Sigmoid layer; (c): the flow chart for the sampling-training iteration.}
        \label{fig:dnn_structure}
    \end{figure}

    The data sampling strategy plays an essential role in algorithm implementation. The dataset size needs to be large enough to train the DNN accurately. However, it is seen that the exhaustive search of the entire input space is not feasible. For the baseline detailed mechanism of 116 species, all possible reduced mechanisms are around $10^{34}$. If labeling each sample costs one millisecond, it will take more than billions of years. As a result, an iterative DNN-assisted sampling method is proposed in this work. The procedure is illustrated in Figure \ref{fig:dnn_structure}c. In the beginning, a few samples are generated randomly with the sparsity $k_0$, which means removing $k_0$ species randomly from the detailed mechanism. The random deleting species surely cause significant errors in some reduced mechanisms. Consequently, part of the sampled reduced mechanisms are labeled as bad reduction ones. More specifically, errors of the benchmark quantities of sampled reduced mechanisms are calculated by Cantera. The problematic reduction leads to large errors in the label. The reduced mechanisms and their errors in benchmark quantities are assembled into the dataset. The DNN training and optimization are performed accordingly. Then, new samples with higher sparsity are generated and merged into the existing dataset. 
    
    The next step is the key for reduced mechanism sampling: using the under-training DNN to evaluate new samples instead of randomly deleting species. Since the sample sparsity grows incrementally, DNN is expected to have the ability to predict the newly sampled mechanisms fast and accurately. DNN helps select only the promising reduced mechanisms to be labeled, discussed in detail in the next section.

\section{Results and discussion} \addvspace{10pt}
    \subsection{Sampled data distribution} \addvspace{10pt}
        The current work proposes a DNN-filter data sampling strategy. The goal is to select promising reduced mechanism samples for DNN training. Therefore, the key question is how to define 'promising' for the selection. First of all, the promising candidates can ignite. Moreover, the reduced mechanism prediction difference compared with the detailed mechanism should be relatively small. For example, the ignition delay time of the detailed mechanism and the reduced mechanism are $\tau_{detailed}$, $\tau_{reduced}$, respectively. The difference should be within two orders of magnitude, i.e., $0.01<\tau_{reduced}/\tau_{detailed}<100$. 
        
        In the beginning, 8192 reduced mechanisms are randomly generated by deleting eight species. The benchmark quantities, such as ignition delay times, equilibrium temperatures, are calculated to label the sampled mechanisms. The resulting dataset is adopted to train the DNN.
        
        In the training-sampling iteration, the sparsity of the sampled mechanism keeps increasing. Higher sparsity indicates a higher possibility that the randomly selected reduced mechanism deteriorate. Consequently, the under-training DNN is utilized to filter the new reduced mechanisms. The number of newly sampled reduced mechanisms is about $16384$ for each iteration. 
        In the current work, the sparsity increases by two for each iterative step. The choice of the sparsity step size is mainly due to the prediction ability of the DNN under training.
        Because a too large step size may lead to a large difference between the datasets of two consecutive iterations, it will potentially cause a large DNN prediction error. 

\begin{figure}[h!]
        \centering
        \includegraphics[width=192pt]{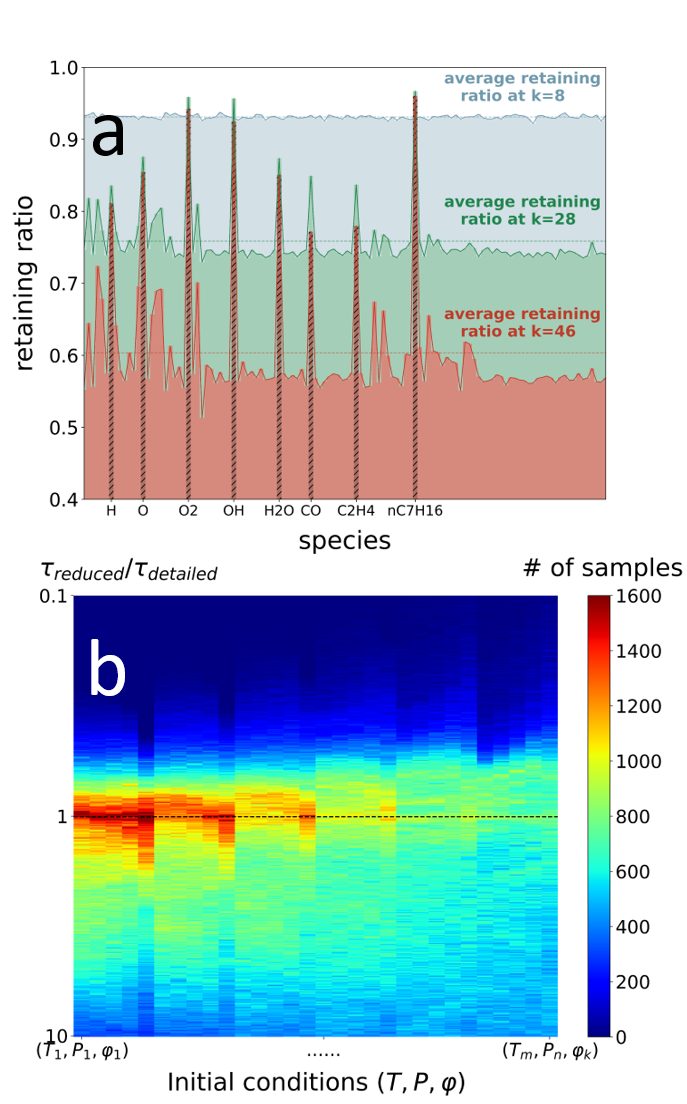}
        \caption{(a): histogram of the retaining ratio for selected reduced mechanisms with different sparsities, the x-axis is the species index, the y-axis is the retaining ratio of species; (b): sampled data distribution. The x-axis represents the initial condition. The y-axis is the ignition delay time ratio between the reduced and detailed mechanisms.}
        \label{fig:sample_data}
        \end{figure}

        Figure \ref{fig:sample_data}a shows sampled data distribution and how it evolves with the iteration. The retaining ratio for the $i$-th species is defined as the number of reduced mechanisms containing the $i$-th species divided by the number of all sampled mechanisms. The dotted line is the averaged retaining ratio. At the beginning of the iteration, at sparsity $k_0=8$ (the blue histogram), all species have a similar retaining ratio due to random sampling. As the training and sampling proceed (the green and red histograms), the retaining ratio significantly differs for different species. The DNN tends to keep important species; thus, these species have higher chances to be selected and higher retaining ratios, such as n-heptane, $O_2$, $CO_2$, and $H_2O$. 
        
        Figure \ref{fig:sample_data}b shows the sampled mechanism distribution of the benchmark quantity of the ignition delay time. The x-axis indexes 30 different initial conditions. The y-axis is the ratio between the ignition delay times of the reduced mechanisms and the detailed mechanism. The number of sampled mechanisms dyes the color. It shows that most of the sampled mechanisms (red and green parts) are located near the centerline $y = 1$. In other words, most of the reduced mechanisms have ignition delay times close to the correct one. It is also important to notice the wide error distribution along the y-direction. It implies that a relatively small part of the reduced mechanisms in the dataset have ignition delay times significantly different from the result of the detailed mechanism, which helps the DNN learn the wrong patterns to avoid similar problematic reductions. This current sample-training iteration proves to be highly efficient. In total, only around $10^5$ samples out of $10^{34}$ all possible samples are selected.

    \subsection{DNN training and prediction performance}
        At the end of the sampling-training iteration, all labeled reduced mechanisms are shuffled and assembled into a comprehensive dataset. $80\%$ of the dataset is used for the DNN training, while the other is the test dataset. The batch size is 128. The initial learning rate is set to 0.01, which decays by $3\%\sim 10\%$ every ten epochs. The training is considered saturated when the test error increases, so-called early stopping.
        
\begin{figure}[h!]
        \centering
        \includegraphics[width=192pt]{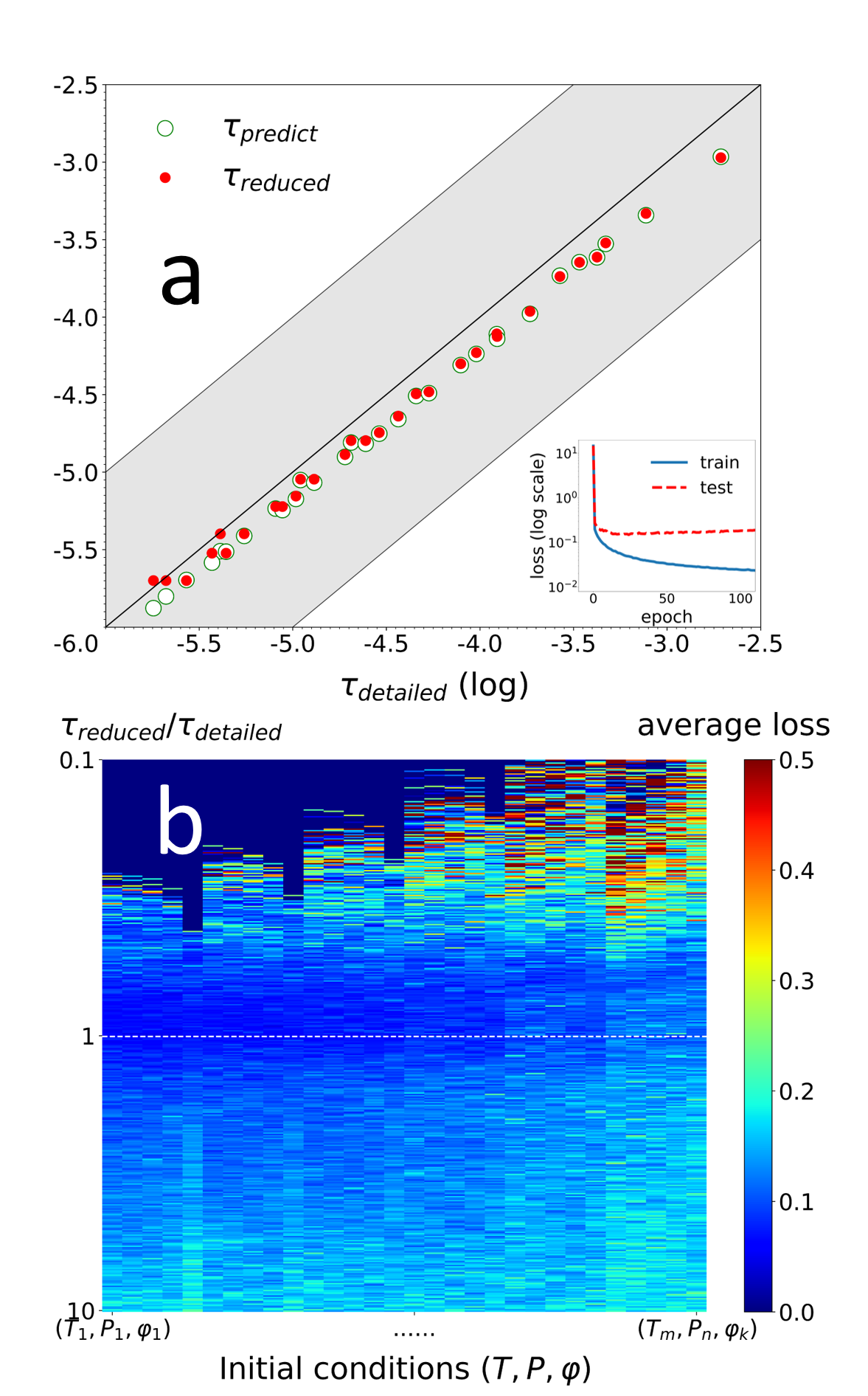}
        \caption{(a) the comparison between DNN prediction and Cantera results of a reduced mechanism; inset: a typical DNN training process; (b) DNN prediction error on the test dataset. The x- and y-axis are the same with Figure 3b: initial conditions and ratio between reduced and detailed mechanism.}
        \label{fig:training}
        \end{figure}

        Figure \ref{fig:training}a shows the DNN prediction for a randomly selected test sample. The x-axis is the ignition delay time of the detailed mechanism calculated by Cantera. The y-axis is the ignition delay time by the DNN prediction (green circle) and Cantera (red dot) for a reduced mechanism. The selected mechanism from the test dataset shows a large deviation compared with the detailed mechanism. However, this deviation is predicted by the DNN accurately. A typical DNN training process is shown in the inset of Figure \ref{fig:training}a.  Figure \ref{fig:training}b shows the DNN prediction error distribution on test samples. The axes are the same as Figure \ref{fig:sample_data}b: the x-axis is the initial condition, the y-axis is the ignition delay time error between the reduced and detailed mechanisms. The DNN prediction error determines the color. Generally speaking, the DNN achieves reasonable accuracy in predicting ignition delay times of reduced mechanisms in different initial conditions. Most of the prediction error is rather small, consistent with the observation in Figure \ref{fig:training}a. Since the samples shown in Figure \ref{fig:training}b are excluded from the DNN training, the good performance on the unseen data demonstrates the reliability of the DNN, which is the prerequisite to obtain a desirable reduced mechanism from the optimization problem. 
        
    \subsection{Inverse optimization problem}
        Optimal reduced mechanisms are achieved by solving inverse optimization problems on the DNN input. Given the DNN function, the optimization goal is to locate the minimum point by SGD. The first step is to generate a random vector, denoted as $\vx_0$. Since we represent a reduced mechanism by a Boolean vector consisting of zero and one, we expect that the inverse problem can produce results similar to Boolean vector.
        Therefore, an entry-wise Sigmoid transformation $\bar{\vx}=\Sigmoid\left(10\vx\right)$ is adopted so that each entry of the solution obtained by the inverse optimization problem is close to zero or one. 
        
        Suppose the benchmark quantities are $P_i$'s, the objective function $F(\vx)$ can be written as: 
        \begin{align}
            F(\vx) &=\sum_{i}{{\lambda_i(P}_i-\text{DNN}_i(\Sigmoid(10\vx))^{2}}\nonumber\\
            &+\lambda_s{||\Sigmoid(10\vx)||}_{L1}
        \end{align}
        where $\lambda_i$ is the weight for the benchmark quantities, $\lambda_s$ is the weight for the sparsity, $\text{DNN}_i$ is the $i$-th output of the DNN. In experiments, we use the same value for $\lambda$'s of the benchmark quantities of the same type, such as $100$ for ignition delay times, $10$ for temperatures, $0.1$ for sparsity weight. Note that $L0$-norm is difficult to optimize, it is a conventional approach to use $L1$-norm in optimization. The input vector is updated as:
        $$\vx_{k+1}=\vx_k-lr\ast{\nabla_{\vx_k}\mathrm{F} \left(\vx_k\right)},$$
        where $lr$ is the learning rate.
    
    
    \subsection{Reduced mechanisms and validation} 
\begin{figure}[h!]
        \centering
        \includegraphics[width=192pt]{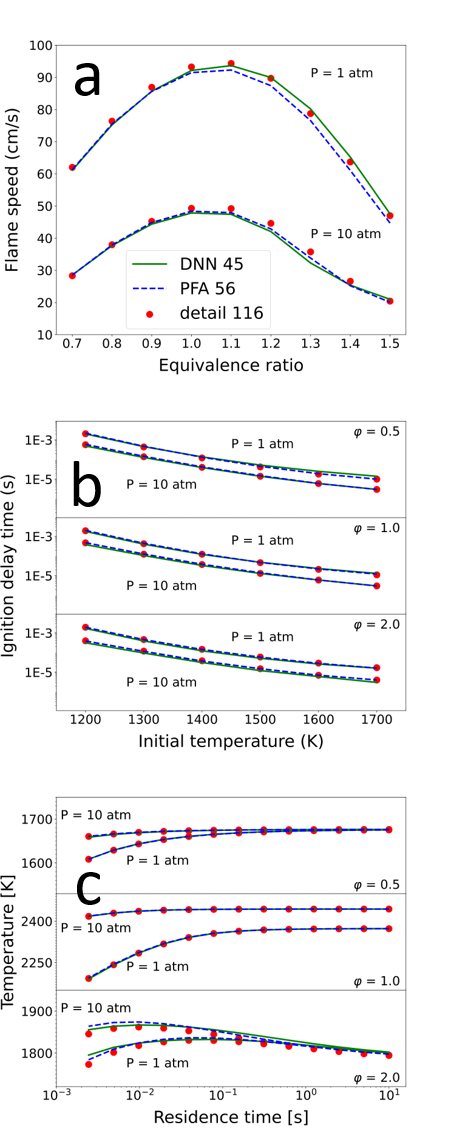}
        \caption{Reduced mechanism (45 species) performance compared with the PFA result and the detailed mechanism. (a): laminar flame speed comparison at P=1 and 10 atm, equivalence ratio $\phi$ = 0.7$\sim$1.5, initial temperature 500K; (b): ignition delay time at P=1 and 10 atm, equivalence ratio $\phi$ = 0.5, 1.0, 2.0, temperature T = $1200\sim1700$ K; (c): temperature under different residence time in PSRs at P=1 and 10 atm, equivalence ratio $\phi$ = 0.5, 1.0, 2.0, initial temperature T = 500K, residence time t = $10^{-3} \sim 10$ s.}
        \label{fig:45red_mech}
        \end{figure}

\addvspace{10pt}
        To demonstrate the validity of the DeePMR, ignition delay time, laminar flame speed, and extinction curves in PSRs are compared using detailed and reduced mechanisms. PFA and the DeePMR generate reduced mechanisms with species number 56 (PFA) and 45 (DNN), respectively. All the calculations are performed by the open-source Python package Cantera. The benchmark tests include different initial conditions from lean to rich mixtures at low and high pressures. The laminar flame speed, ignition delay time, and extinction curves comparison are shown for initial pressure $P=1 {\rm atm}$ and $P=10 {\rm atm}$ in Figures \ref{fig:45red_mech}a, \ref{fig:45red_mech}b, and \ref{fig:45red_mech}c, respectively. In most benchmark tests, reduced mechanisms from the DNN and the PFA produce equally accurate results, the averaged error of ignition delay time, laminar flame speed, and PSR temperature is within $15\%$.
        
\begin{figure}[h!]
        \centering
        \includegraphics[width=192pt]{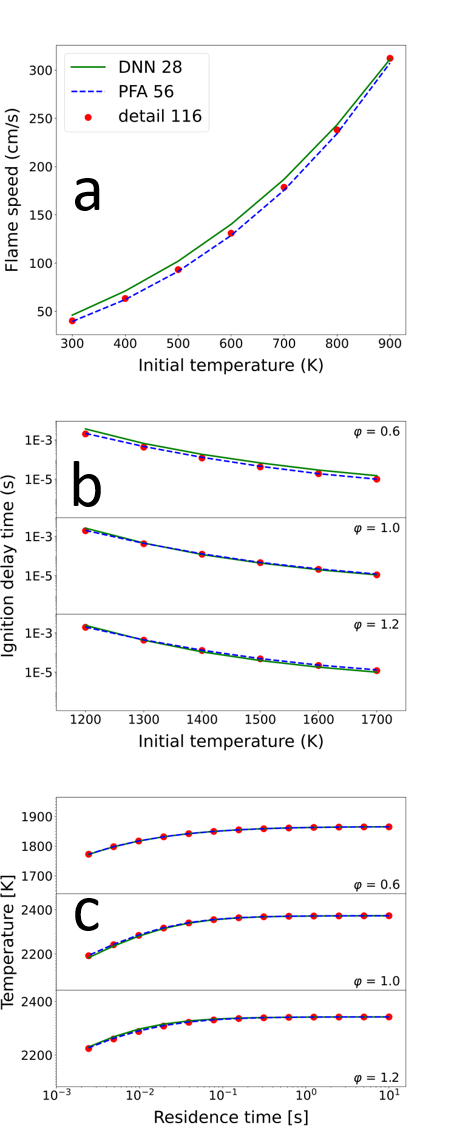}
        \caption{Reduced mechanism (28 species) performance compared with the PFA result and the detailed mechanism. (a): laminar flame speed comparison at P=1 atm, equivalence ratio $\phi$ = 1.0, initial temperature $T=300\sim900$ K; (b): ignition delay time comparison at P=1 atm, equivalence ratio $\phi$ = 0.6, 1.0, 1.2, initial temperature T = $1200\sim1700$ K; (c): temperature comparison under different residence time in PSRs at P=1, equivalence ratio $\phi$ = 0.6, 1.0, 1.2, initial temperature T = 500K, residence time t = $10^{-3} \sim 10$ s.}
        \label{fig:28red_mech}
        \end{figure}

        The DNN method can reduce the detailed mechanism more aggressively if the targeted scenarios are near stoichiometric conditions. Figure \ref{fig:28red_mech} shows the reduced mechanism with 28 species, half of the size of the PFA reduced mechanism. The only difference in performing DeePMR is to remove non-stoichiometric benchmark quantities from the DNN data labeling. The reduced mechanism shows a reasonable agreement with the detailed mechanism at P = 1 atm and equivalence ratio between 0.6 and 1.2.
        
\begin{figure}[ht]
            \centering
            \includegraphics[width=192pt]{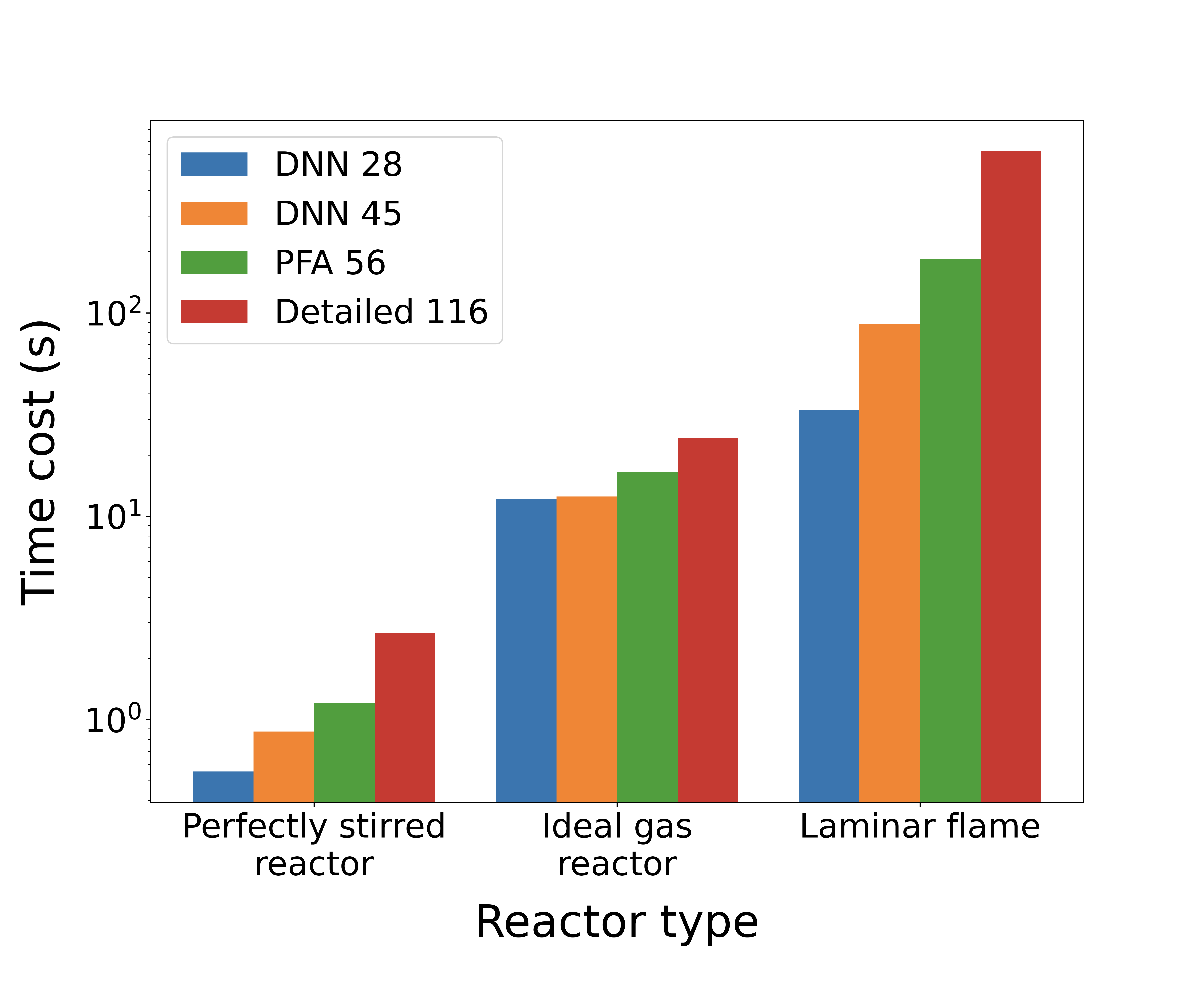}
            \caption{Computation cost comparison among different mechanisms.}
            \label{fig:timecost}
        \end{figure}

       Figure \ref{fig:timecost} shows the CPU time cost comparison using different mechanisms in perfectly stirred reactor, ideal gas reactor and laminar flame simulations. The results demonstrate that the reduced mechanism significantly accelerates various simulations. The reduced mechanism by DNN with 45 species have an averaged 3x speedup factor compared with the detailed mechanism, and 30\% efficiency improvement compared with the reduced mechanism with 56 species from PFA method. The reduced mechanism with 28 species save additional 50\% CPU time compared with the 45 species mechanism.  
        
\section{Conclusions} \addvspace{10pt}
   The current paper proposes a deep learning-based  model  reduction (DeePMR) method for simplifying combustion chemistry mechanisms. The key idea is to use an end-to-end DNN to predict a reduced mechanism accuracy. The input is the reduced mechanism, and the output is the selected benchmark quantities. The benchmark quantities can be ignition delay time, equilibrium temperature, laminar flame speed, depending on the scenario for the reduced mechanism.
   
    The sampling is the major challenge due to the extremely large number of possible reductions. Consequently, an iterative procedure of sampling and DNN training is adopted. The iteration takes advantage of the DNN prediction to filter reduced mechanisms. Only promising candidates are retained and labeled. The sampling-training iteration generates a representative dataset efficiently.
    
    The final reduced mechanism is obtained by solving an inverse optimization problem. The optimization target is to simultaneously minimize the reduced mechanism's error and size, where the well-trained DNN provides the objective function. 

    The results show that the DNN reduced mechanisms have a smaller size and comparable accuracy than the path flux analysis method, which demonstrates the great potential for machine-learning methods to perform model reduction for combustion kinetics.
\acknowledgement{Acknowledgments} \addvspace{10pt}
This work is supported by the Shanghai Sailing Program, the Natural Science Foundation of Shanghai Grant No. 20ZR1429000  (Z. X.), the National Natural Science Foundation of China Grant No. 62002221 (Z. X.), the National Natural Science Foundation of China Grant No. 12101402 (Y. Z.), Shanghai Municipal of Science and Technology Project Grant No. 20JC1419500 (Y.Z.), Shanghai Municipal of Science and Technology Major Project No. 2021SHZDZX0102 (Z. X., Y.Z.), and the HPC of School of Mathematical Sciences (Z. X., Y.Z.) and the Student Innovation Center (Z. X., Y.Z.) at Shanghai Jiao Tong University, and AI for Science Institute, Beijing.

 \footnotesize
 \baselineskip 9pt


\bibliographystyle{pci}
\bibliography{library}


\newpage

\small
\baselineskip 10pt



\end{document}